\documentclass[10pt,twocolumn,letterpaper]{article}

\usepackage{cvpr}
\usepackage{times}
\usepackage{epsfig}
\usepackage{graphicx}
\usepackage{amsmath}
\usepackage{amssymb}

\usepackage{subfigure}
\usepackage{booktabs}
\usepackage{epstopdf}
\usepackage{algorithmic}
\usepackage{algorithm}
\usepackage{lipsum}
\newcommand\blfootnote[1]{%
\begingroup
\renewcommand\thefootnote{}\footnote{#1}%
\addtocounter{footnote}{-1}%
\endgroup
}

\cvprfinalcopy

\begin{document}

\title{HAMBox: Delving into Online High-quality Anchors Mining \\
for Detecting Outer Faces}

\author{
Yang Liu$^{1\dagger *}$, Xu Tang$^{1\dagger}$, Xiang Wu$^{2}$, Junyu Han$^{1}$, Jingtuo Liu$^{1}$, Errui Ding$^{1}$ \\
$^1$\emph{Department of Computer Vision Technology (VIS), Baidu Inc.} \\
$^2$\emph{National Laboratory of Pattern Recognition, Institute of Automation, Chinese Academy of Sciences.} \\
\{v\underline{ }liuyang45, tangxu02, hanjunyu, liujingtuo, dingerrui\}@baidu.com\; \{alfredxiangwu\}@gmail.com
}

\maketitle

\begin{abstract}
Current face detectors utilize anchors to frame a multi-task learning problem which combines classification and bounding box regression. {Effective anchor design and anchor matching strategy enable face detectors to localize faces under large pose and scale variations.} However, we observe that more than 80\% correctly predicted bounding boxes are regressed from the unmatched anchors {(the IoUs between anchors and target faces are lower than a threshold)} in the inference phase. It indicates that these unmatched anchors perform excellent regression ability, {but the existing methods neglect to learn from them.}
In this paper, we propose an Online High-quality Anchor Mining Strategy (HAMBox), which explicitly helps outer faces compensate {with} high-quality anchors. Our proposed HAMBox method could be a general strategy for anchor-based {single-stage} face detection. Experiments on various datasets, including WIDER FACE, FDDB, AFW and PASCAL Face, demonstrate the superiority of the proposed method.
Furthermore, our team win the championship on the Face Detection test track of WIDER Face and Pedestrian Challenge 2019. {We will release the codes with PaddlePaddle.}
\blfootnote{$^{\dagger}$Equal contribution. $\ ^{*}$Work done during an internship at Baidu VIS.}
\vspace{-0.3cm}
\end{abstract}

\section{Introduction}
Face detection is a fundamental task for many high-level face-based 
applications, such as face alignment \cite{zhang2014coarse}, face recognition \cite{antipov2017face} and face aging \cite{wang2018face}. Deriving from early 
face detectors with hand-crafted features, modern detectors have been {significantly} improved owing to the robust features learnt with deep Convolutional Neural Networks (CNNs) \cite{krizhevsky2012imagenet}. Current state-of-the-art face detectors are usually based on anchor-based deep CNNs, inspired by their successes on the general object detection.

Different from general object detectors, face detectors often face smaller variations of aspect ratios (from 1:1 to 1:1.5) but much larger scale variations (face area, from several pixels to thousands of pixels). Considering the large variations of scales,  Zhang et al. \cite{zhang2017s3fd} tile anchors on a wide range of layers and design anchor scales according to the effective receptive field. Current state-of-the-art detectors \cite{tang2018pyramidbox,li2019dsfd} capture the locations of various face scales by utilizing Feature Pyramid Network (FPN) \cite{lin2017feature}. FPN is an effective way to exploit the inherent multi-scale features for constructing feature pyramids in a top-down manner. It adopts lateral connection from the high-level deeper features to the low-level ones.
Then from the perspective of designing anchor setting, anchor-based detectors with FPN continue to resolve this by raising the number of anchors from different aspects (e.g., anchor stride,  and ratio anchors \cite{zhu2018seeing,wang2017face}). However, increasing the number of anchors remarkably reduces the performance of a face detector, especially when adopting the feature map conv2 or P2 (in Resnet-50) for recalling small faces empirically.

\begin{figure}[t]
    \subfigure[Average Number of Anchors Matched to Each Face ]{
    \includegraphics[width=0.2\textwidth]{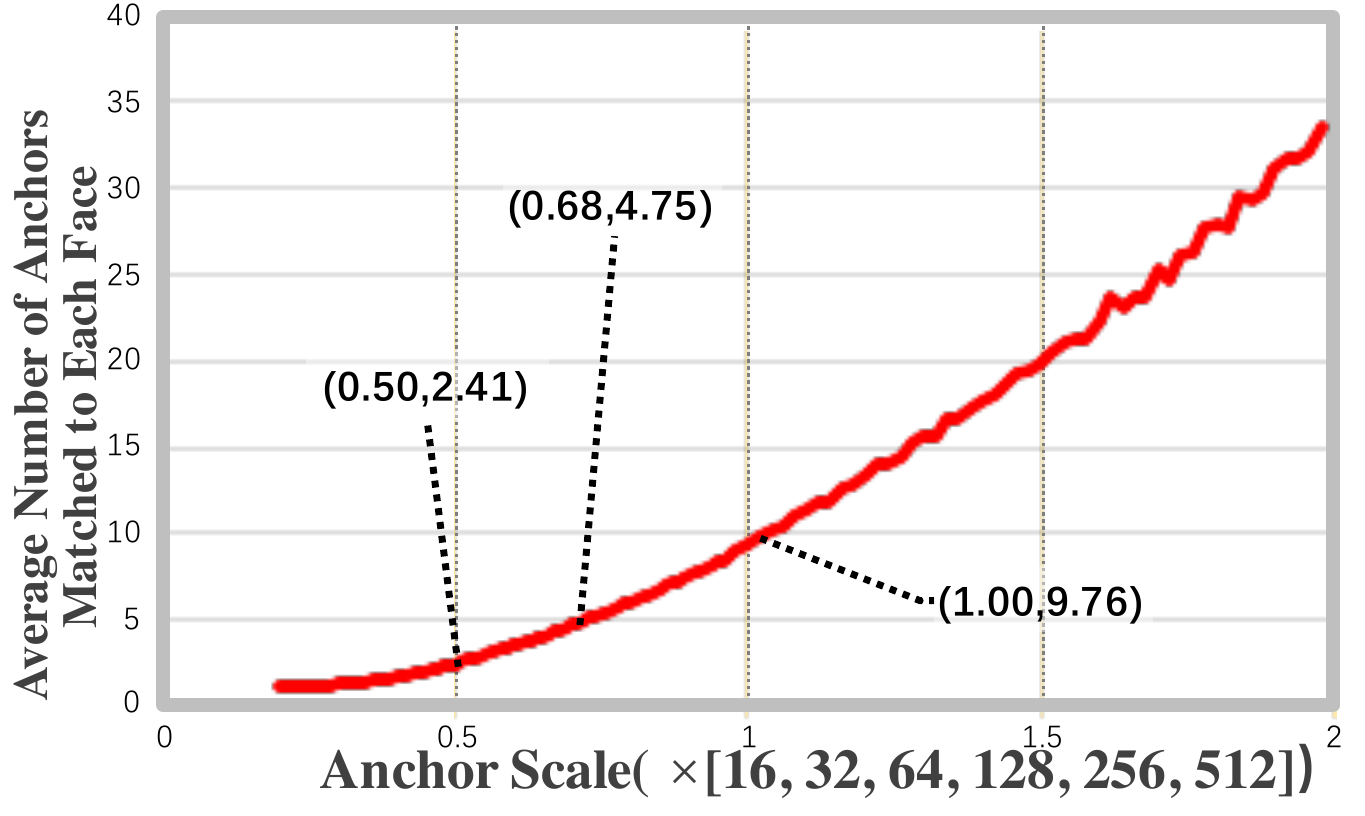}
    \label{img1_a}
    }
    \hspace{0.2in}
    \subfigure[Proportion of Faces that can Match with Anchors]{
    \includegraphics[width=0.2\textwidth]{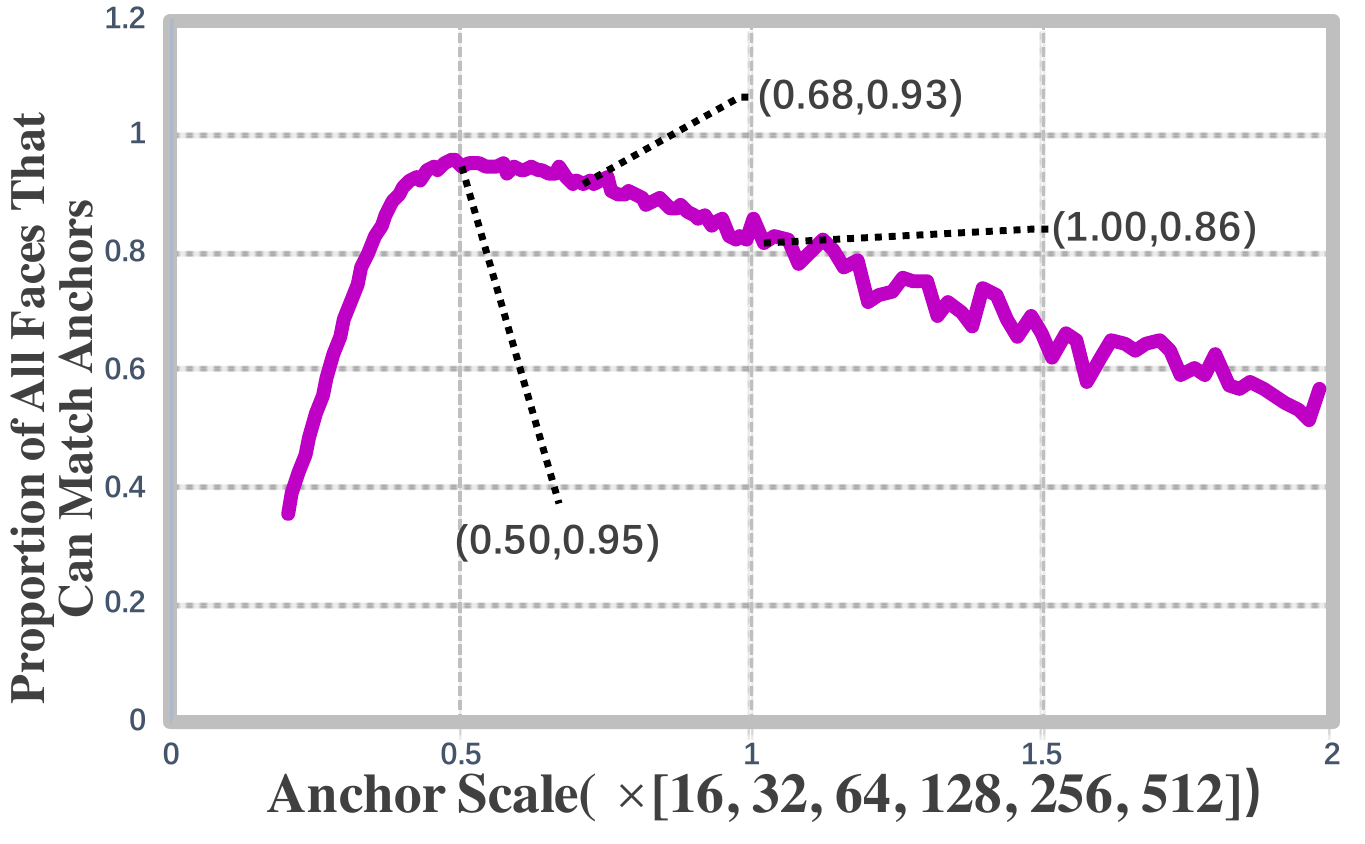}
    \label{img1_b}
    }
    \caption{Two crucial factors in designing anchor scales on the WIDER FACE dataset. (a) As the scale of anchor increases, the average number of anchors matched to each face also increases. (b) The proportion of faces that can match the anchors decreases significantly outside a specific interval ([0.43, 0.7]).}
    \label{img1}
    \vspace{-0.3cm}
\end{figure}

\begin{figure}[t]
    \subfigure[Cumulative Desity Curve of IoU]{
    \includegraphics[width=0.2\textwidth]{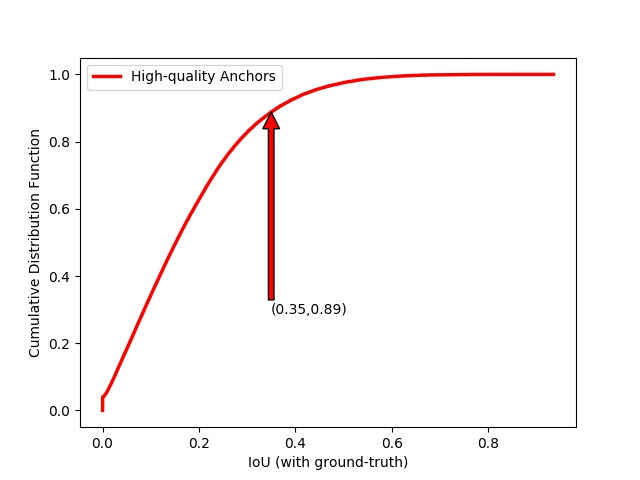}
    \label{img2_a}
    }
    \hspace{0.3in}
    \subfigure[Performance of Compensated Anchors]{
    \includegraphics[width=0.2\textwidth]{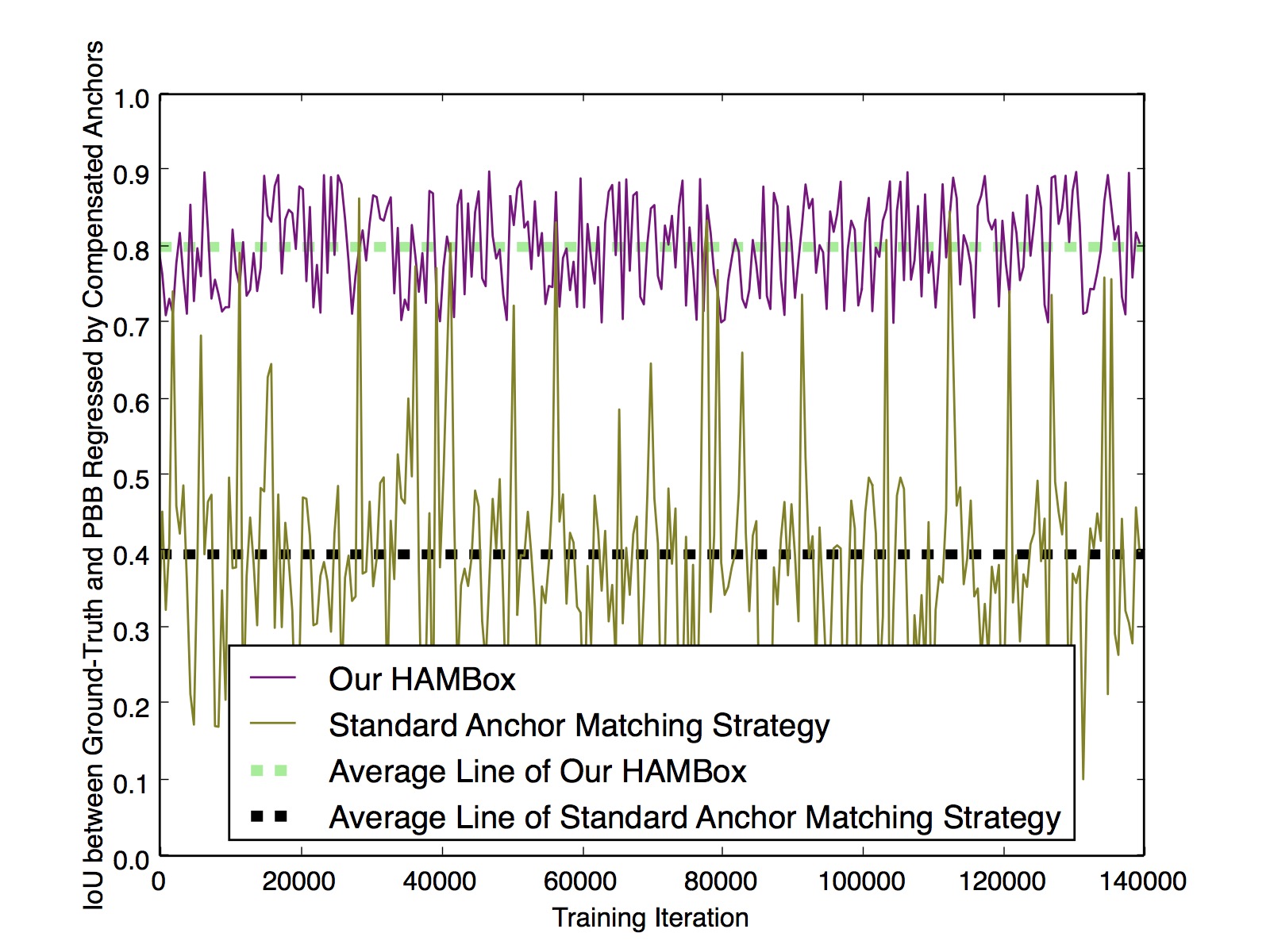}
    \label{img2_b}
    }
    
    \subfigure[Proportion of unmatched High-quality Anchors]{
    \includegraphics[width=0.2\textwidth]{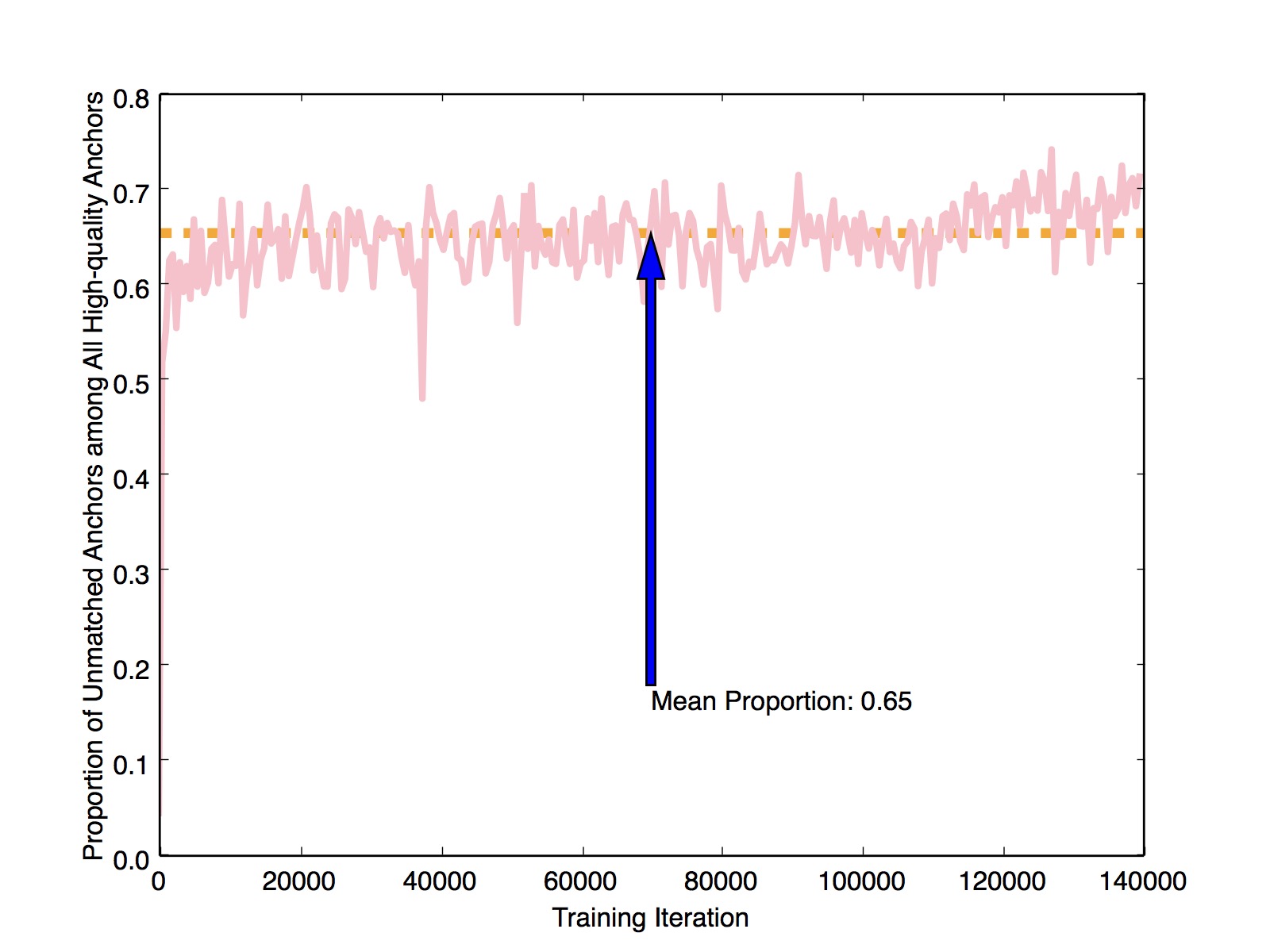}
    \label{img2_c}
    }
    \hspace{0.3in}
    \subfigure[Performance of Matched High-quality Anchors]{
    \includegraphics[width=0.2\textwidth]{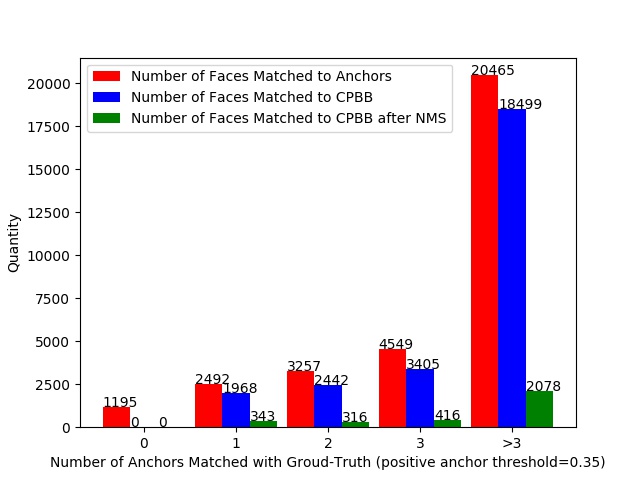}
    \label{img2_d}
    }
	
    \caption{The problem of standard anchor matching strategy during training and inference (on the WIDER FACE dataset). (a) During inference, only 11\% of all correctly predicted bounding boxes are regressed by matched anchors. (b) PBB represents {`Predicted Bounding Boxes'}. When using our HAMBox strategy, the {IoUs between ground-truths and predicted bounding boxes regressed by compensated anchors are} much higher than{standard anchor matching strategy} during training. (c) During training, the average number of unmatched high-quality anchors {occupies a surprisingly} 65\% proportion of all high-quality anchors. (d) CPBB represents {`Correctly Predicted Bounding Boxes'}. During inference, the number of matched high-quality anchors dramatically decreases after NMS, representing some unmatched anchors have higher regression ability. All these results demonstrate that the standard anchor matching strategy  can not utilize {high-quality} negative anchors effectively, which play essential roles whatever during training or inference.}
    \label{img2}
\end{figure}

\begin{figure*}
    \centering
    \includegraphics[width=0.7\textwidth, height=2.1in]{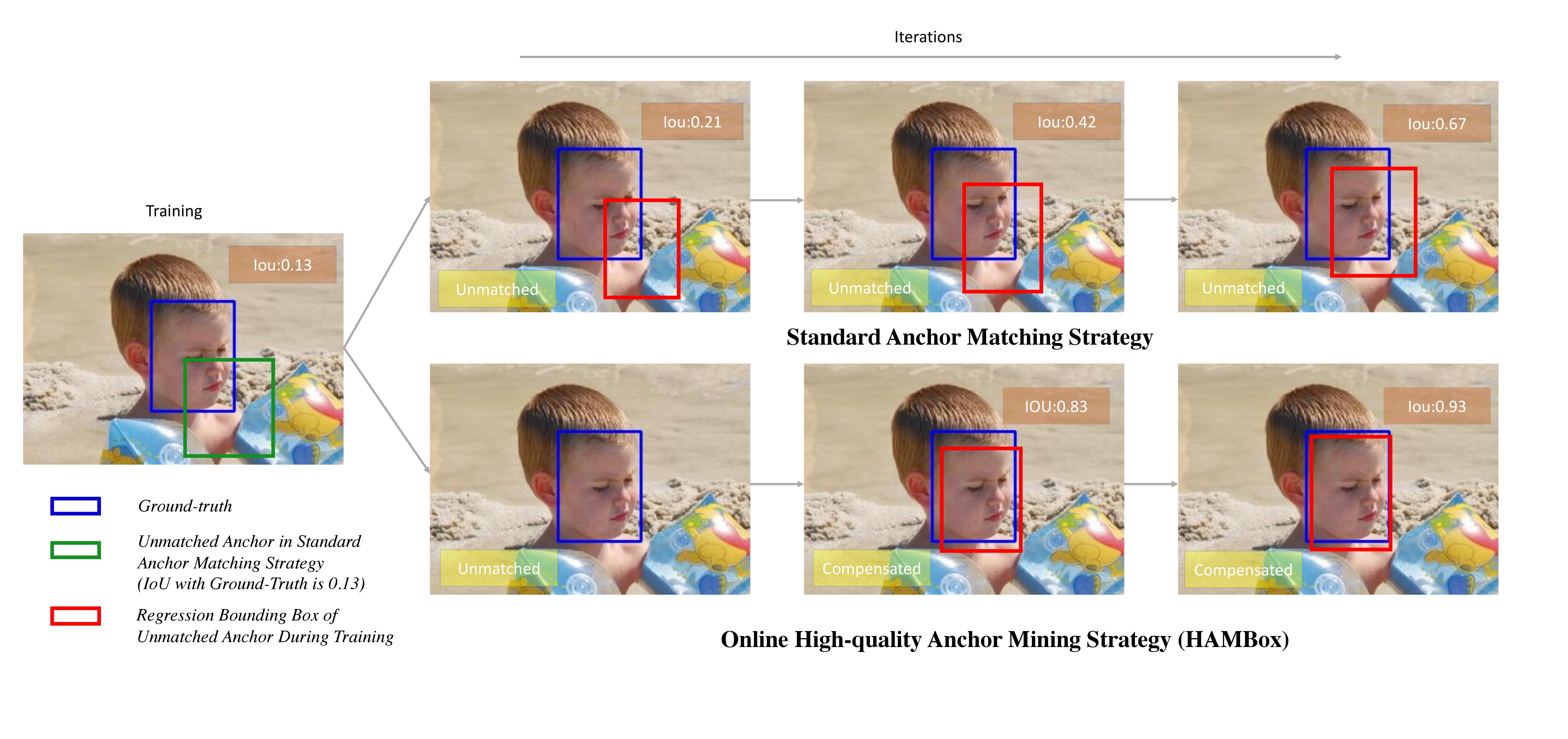}
    \caption{Visualization of the quality of compensated anchors through two methods. In the early stage of training, our method does not compensate anchors for outer faces. Then with the increasing of training iteration, our method is gradually mining unmatched high-quality anchors for outer ones, which have higher IoU than anchors generated by standard anchor matching strategy.}
    \label{img2_e}
\end{figure*}

As far as we know, for an anchor-based detector, effective anchor design strategies are necessary to achieve high performance. S$^3$FD \cite{zhang2017s3fd} adopts single scale and aspect ratio anchors for each detection stage. Nonetheless, choosing the proper anchor scale remains a big challenge, which generally produced by the following misalignment phenomenon.  
Figure \ref{img1} shows `the average number of anchors matched to each face' and `the proportion of all faces that can match the anchors' across different anchor scales, which are two {indicative} factors to be considered in designing proper anchor scale. With the increase of anchor scales, although the number of anchors matched with each face steadily grows,   
the proportion of faces which are capable of matching anchors gradually descends. Moreover, this misalignment usually leads to a heuristical anchor scale designation.
   
To alleviate the imbalance between `the average number of anchors matched to each face' and `the proportion of all faces that can match the anchors' as discussed above, two representative solutions have been proposed: Firstly, S$^3$FD \cite{zhang2017s3fd} introduces an anchor compensation strategy by offsetting anchors for outer faces\footnote{{Faces cannot match enough positive anchors. In our paper, we set the number as hyper-parameter $K$ detailed in the Subsection \ref{online_high_quality}.}}; Secondly, Zhu et al. \cite{zhu2018seeing} formulate a metric named Expected Maximum Overlap (EMO) to obtain more reasonable anchor stride and receptive field. All these solutions focus on helping outer faces match more anchors during the training phase. However, they also bring a large number of redundant or low-quality anchors. (see the olive line of Figure \ref{img2_b}). 

In this paper, we conduct an anchor matching statistic {experiment} on a well-trained face detector \cite{tang2018pyramidbox,li2019progressively} and find an intriguing phenomenon. The red line in Figure \ref{img2_a} represents the cumulative distribution curve of IoU between the ground-truth and the anchors which can be regressed to correctly predicted bounding boxes. We surprisingly observe that only 11\% of all correctly predicted bounding boxes are regressed by matched anchors. So, not only the matched anchors but also some unmatched ones
play a critical role in face detection. However, in the phase of training, those unmatched anchors
are assigned with background labels, which are unreasonable supervision signals for classification branch consequently. Effectively leveraging these unmatched anchors is expected to improve the detection performance.

Motivated by this observation, we identify two key issues in current anchor matching strategies as follows: 
\begin{itemize}
	\item \textbf{The majority of compensated anchors are of low-quality.} Figure \ref{img2_b} shows the regression ability of compensated anchors during training when adopting the standard anchor matching strategy \cite{ren2015faster}. Apparently, compensated anchors have a poor performance on location regression (average IoU between the bounding boxes regressed by compensated anchors and the ground-truth is 0.42). In other words, this method helps those outer faces matching more low-quality anchors, instead of high-quality ones.
	\item \textbf{Many unmatched anchors in the training phase actually have strong localization ability}. As shown in Figure \ref{img2_c}, around 65\% of all high-quality anchors\footnote{The intersection-over-union (IoU) between its regression bounding box and corresponding ground-truth is higher than 0.5.} are unmatched anchors during training. Based on the above observations, we argue that the current anchor matching strategy is neither flexible nor sufficient to utilize the anchors  in face detection.
 As illustrated in Figure \ref{img2_d}, the red, blue and green bars denote the number of faces matched to anchors (IoU$\textgreater$0.35\footnote{This denotes the IoU between anchor and target face in the training phase.}), matched to correctly predicted bounding boxes (IoU$\textgreater$0.5\footnote{This denotes the IoU between the predicted bounding box of matched anchor and the target face.}) and matched to correctly predicted bounding boxes after Non-Maximum Suppression (NMS). It is obvious that the correctly predicted bounding boxes regressed by unmatched anchors suppress the ones regressed by matched anchors during NMS. Lots of unmatched anchors also have strong abilities for regression. 
\end{itemize}

To address this issue, we propose an Online High-quality Anchor Mining Strategy (HAMBox) method. The idea is to mine those high-quality anchors consistently to help outer faces compensate more anchors with the ability of precise regression. Figure \ref{img2_b} and Figure \ref{img2_e} show that the quality of our compensated anchors has a  significant enhancement than standard anchor mathing strategy's. In Figure \ref{img2_e}, when using standard anchor matching strategy, the unmatched anchors {are} assigned with background labels. With the increase of training iteration, our Online High-quality Anchor Compensation Strategy is gradually mining unmatched high-quality anchors for outer faces. Moreover, the unmatched anchors could regress high-quality bounding {boxes} with higher IoU than anchors generated by standard anchor matching strategy. After mining high-quality anchors, we further propose regression-aware focal loss to {effectively weight} those new compensated high-quality anchors. Dynamic weights based on IoU are added for new compensated anchors mainly by considering the weak connection between location and classification. Benefiting from online high-quality anchor compensation strategy and regression-aware focal loss, we achieve 91.6\% AP on the  WIDER FACE \cite{yang2016wider} validation hard set, with the baseline of RetinaNet \cite{lin2017focal}. Furthermore,  we add some popular modules, including SSH head \cite{najibi2017ssh}, deep head \cite{lin2017focal}, and pyramid anchors \cite{tang2018pyramidbox}, and achieve 93.3\% AP, which outperform current state-of-the-art model \cite{li2019dsfd} by a large margin of 2.9\% AP.

In summary, our main contributions {can be summarized as}: 
\vspace{-0.3cm}
\begin{itemize}
    \item We observe an inspiring phenomenon that some unmatched anchors have strong regression ability, and the current box regression branch neglects to learn unmatched anchors.
   	\vspace{-0.2cm}
    \item Based on the observations, we propose an Online High-quality Anchor Mining Strategy (HAMBox) to sample high-quality anchors for training. Benefiting from HAMBox, we can provide sufficient and effective anchors for outer faces in the training phase;
    \vspace{-0.2cm}
    \item Thanks to the high-quality anchors, a regression-aware focal loss assists in face detector training with a flexible way;
    \vspace{-0.2cm}
    \item Our approach outperforms the state-of-the-art methods by 2.9\% and 2.3\% AP on the WIDER FACE validation and test hard-set, respectively. Moreover, we achieve 57.45\% (validation) / 57.13\% (test) mAP on the Face Detection track of WIDER Face and Pedestrian Challenge 2019.
\end{itemize}

\section{Related Work}
Face detection is a fundamental yet challenging computer vision task. Viola and Jones \cite{viola2004robust} first utilizes Haar features and AdaBoost to train a face detector. After that, more following works pay attention to combining multi models to get discriminative features. For example, DPM \cite{felzenszwalb2009object} proposes an extra model to capture human lateral feature and merges it with front and back body features. All the face detectors based on hand-craft features are optimized with each sub-model separately. Due to both weak features and classifiers, the performance of these face detectors {is} limited in the practical scenario.

Recently, owing to the rapid development of deep convolutional networks \cite{krizhevsky2012imagenet,he2016deep,simonyan2014very,szegedy2015going} on image classification and object detection, face detection has made significant progress on large variations, including poses, scales, blur and occlusions, etc., in practice. By introducing the core ideas of hand-craft face detector, Cascade CNN and Multi-task CNN (MTCNN) propose a coarse-to-fine framework to capture faces via deep CNNs. With the flourish of general object detectors \cite{ren2015faster,liu2016ssd,chen2019dubox}, \cite{jiang2017face} and SSH \cite{najibi2017ssh} introduce anchor-based detectors to face detection. Yang et al. \cite{yang2016wider} collect WIDER FACE dataset, which contains rich annotations, including occlusions, poses, event categories, and face bounding boxes. The WIDER FACE dataset pushes forward to the research of face detection, focusing on the extreme variations, including scale, pose and occlusion. Recently, most state-of-the-art face detectors focus on these extreme variations with three following aspects: image pyramid, feature pyramid and context module. HR \cite{hu2017finding} designs image pyramids of the low, medium and high resolutions for training and testing, which significantly boosts the performance on extreme scale variations (from several {pixels to thousands of pixels}). FAN \cite{wang2017face} introduces attention modules and feature pyramid network \cite{lin2017feature} to capture occluded faces. SSH \cite{najibi2017ssh} builds a detection module with a rich receptive field. PyramidBox \cite{tang2018pyramidbox} formulates a data-anchor-sampling strategy to increase the proportion of small faces in the training data. Moreover, by designing a scale propose network, SAFD \cite{hao2017scale} generates a scale histogram and further automatically normalizes face scales prior for optimizing face detectors. DSFD \cite{li2019dsfd} introduces small faces supervision signals on the backbone, which implicitly {boosts} the performance of pyramid features.

Considering some works on anchor design and sampling strategies, S$^3$FD \cite{zhang2017s3fd} proposes a new anchor matching strategy which helps the outer faces match more anchors. SRN \cite{chi2019selective} introduces a Selective Two-step Classification to ignore training easy sample anchors in the second stage. ZCC \cite{zhu2018seeing} introduces Expected Max Score to evaluate the quality of anchor matching, which helps to design anchor stride. Group sampling \cite{ming2019group} conducts lots of experiments on the ratio of matched and unmatched anchors, which emphasizes the importance of the ratio for matched and unmatched anchors. In this paper, inspired by the anchor matching strategy in S$^3$FD \cite{zhang2017s3fd} and the statistical curve discussed  in Figure \ref{img1} and Figure \ref{img2}, we propose {an} Online High-quality Anchor Mining Strategy (HAMBox), as well as a regression aware focal loss. Benefiting from these methods, we achieve a strong face detector, compared with other state-of-the-art face detection methods.

\begin{figure*}
    \centering
    \subfigure[Face Matched to Anchor on the First Step of Standard Anchor Matching Strategy]{
    \includegraphics[width=1.5in, height = 1.2in]{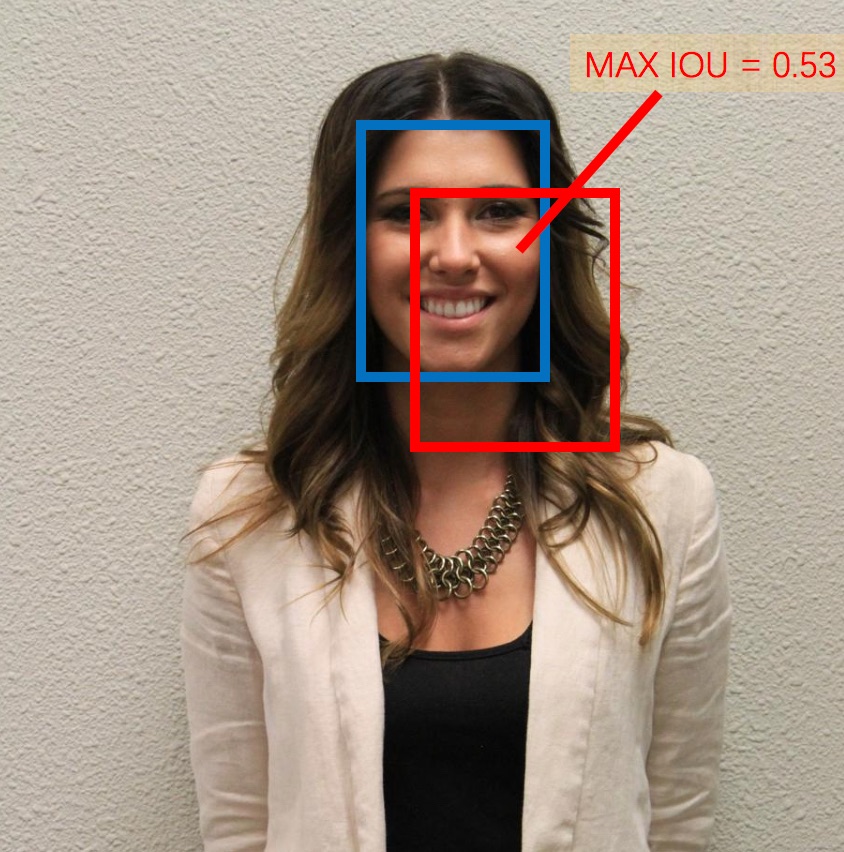}
    \label{img3_a}
    }
    \hspace{3ex} 
    \subfigure[Face Matched to Anchor on the Second Step of Standard Anchor Matching Strategy]{
    \includegraphics[width=1.5in, height = 1.2in]{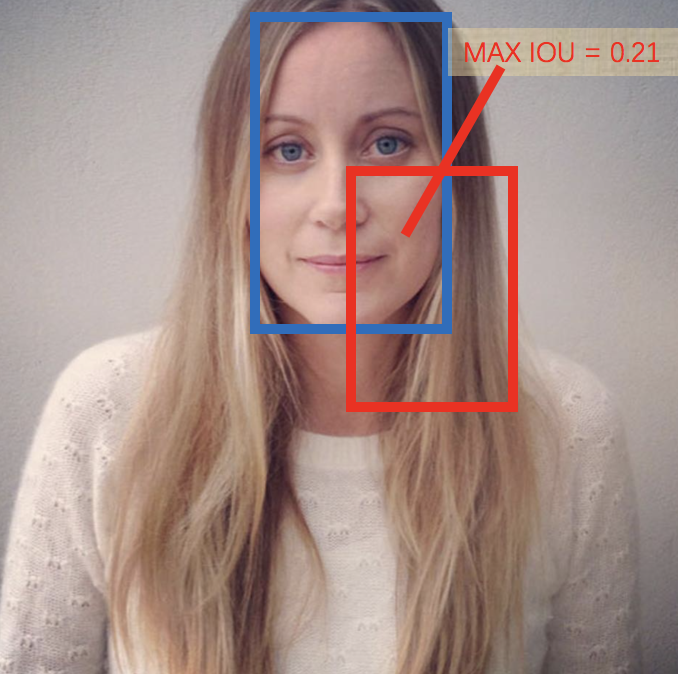}
    \label{img3_b}
    }
    \hspace{3ex}
    \subfigure[Cummulative Density Curve of IoU]{
    \includegraphics[width=2in, height = 1.2in]{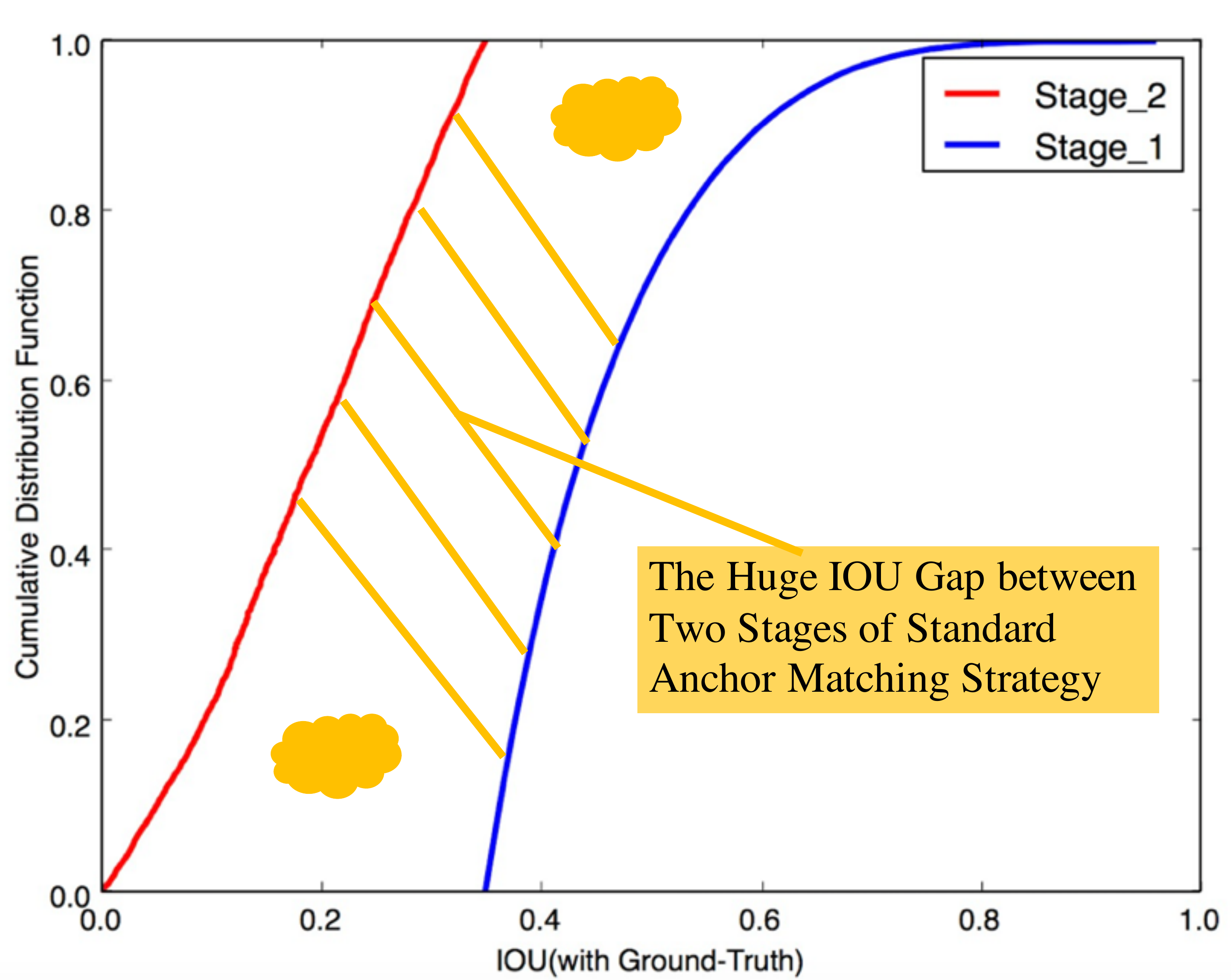}
    \label{img3_c}
    }
    \caption{(a) (b) Two different stages on standard anchor matching strategy, the blue rectangle represents ground-truth and the red one is an anchor matched with it. (c) Cumulative Density Curve of IoU between ground-truth and its matched anchor on different stages.}
    \label{img3}
\end{figure*}

\section{Online High-quality Anchor Mining}
This section presents the proposed Online High-quality Anchor Mining Strategy (HAMBox) to compensate outer faces with the most proper anchors.
We firstly build our {high-recall} face detector based on RetinaNet \cite{lin2017focal}. Then we demonstrate the online high-quality anchor compensation strategy in detail. Finally, we {formulate} a regression-aware focal loss for the compensated anchors.

\subsection{High-recall Anchor-based Face Detector}
\label{high-recall anchor-based detector}
Current anchor-based face detectors utilize predefined anchors to frame a multi-task learning problem, which combines classification and bounding box regression branches. We start with RetinaNet \cite{lin2017focal} as the baseline. The backbone is ResNet-50. Following the settings in \cite{zhang2017s3fd}, we employ the feature map of conv2 layer to improve the performance of face detector. The reason is that around 40\% faces are matched to conv2 anchors on the WIDER FACE benchmark. Furthermore, it is important to design anchors for training a well-performed detector. Therefore, different from the general object detection with multiple anchor scales and aspect ratios, we set only one anchor scale and one aspect ratio at each prediction layer for our default anchor settings.

Inspired by statistical results in Figure \ref{img1}, we change the anchor scale\footnote{In our method, anchor scale is set to 0.68$\times$\{16, 32, 64, 128, 256, 512\} and ratio is 1:1 at different prediction layers.} to match more extreme face scales. The advantage and disadvantage of this anchor setting are equally obvious. From the perspective of advantage, our strategy can match over 95\% of {all the} faces on the WIDER FACE benchmark, a small difference comparing to multi-scale and ratio anchors that can match 98.46\% of faces. At the same time, our method uses three times or nine times fewer anchors than the latter anchor setting with multi-scale and ratio, leading model to focus more on the regression of useful anchors and further get higher detecting performance. From the perspective of disadvantage, it is harmful to the robustness of the model because decreasing the number of faces matched to anchors. This obstruction will be resolved in the following two sections.

\subsection{Online High-quality Anchor Compensation Strategy}
\label{online_high_quality}

After finishing the design of anchor scale and ratio, we further need to allocate anchors with their nearest adjacent ground-truth or background. As shown in Figure \ref{img3},
{the} current anchor matching strategy consists of two steps. A face firstly matches anchors with IoU higher than a threshold. Then {faces that do not match} with any anchor would be compensated with anchors that have the max IoU with them. Obviously, compensated anchors in the second step may reduce the performance of regression and classification of the network since these anchors initially have
lower IoU with faces, as shown in Figure \ref{img3_c}.

In Figure \ref{img2_b}, we surprisingly find that with the increase of iterations, some unmatched anchors have the ability {to make correct predictions} while those are ignored on regression branch and even assigned as background on classification branch. Inspired by this observation, we propose an Online High-quality Anchor Compensation strategy to resolve current misaligned supervision signal. 
Firstly, 
each face  matches the anchors with IoU higher than a threshold, but for those remaining outer faces, we do not compensate any anchors. Secondly, at the end of forward propagation during training, each anchor computes  regression bounding box through its related regression coordinates. 
We define this regression bounding box as $B_{reg}$ and $F_{outer}$ represents outer faces. Finally, for each face in $F_{outer}$, we compute its IoU with $B_{reg}$ and compensate this face with $N$ extra unmatched anchors. We define all IoUs as $IoU_{set}$. These $N$ compensated anchors are selected according to two rules. 1) The IoUs between their corresponding regression bounding boxes and target faces should be greater than $T$ ($T$ represents an online positive anchor threshold). 2) These IoUs (calculated in rule 1) should be in the top-$K$ highest IoU in $IoU_{set}$. $K$ is a hyperparameter that represents the max number of anchors that $F_{outer}$ can be matched with. If $N$ is greater than $K-M$ after filtering out by above two rules, we select top-($K-M$) highest IoU anchors in these $N$ unmatched anchors to compensate this face and set $N=K-M$. $M$ denotes the number of anchors that faces already matched with in the first step. We have done {many} experiments in ablation study by varying $T$, $K$.
Details can be seen in Algorithm \ref{algorithm_1}.

\subsection{Regression-aware Focal Loss}
 After the analysis of two subsections above, we have 
mined those high-quality anchors and the following problem is 
 to make full use of these anchors effectively. 
 Furthermore, we propose a regression-aware focal loss to give more reasonable weights to those new compensated high-quality anchors, which are newly mined for outer faces by Online High-quality Anchor  Compensation Strategy. 
 
    Two improvements have been made on focal loss \cite{lin2017focal}. (1) Considering the weak
    connection between location and classification on new compensated anchors, we add dynamic weights based on IoU for these compensated ones. (2) We define anchors satisfying the following three conditions simultaneously as ignored anchors (which are not optimized during training): a) Belong to the high-quality anchors. b) Be assigned as background in the first step of anchor matching strategy. c) {Not included in} new compensated anchors. We define the loss as:
\begin{equation}
\resizebox{0.43\textwidth}{!}{$
\begin{array}{ll}
&L_{cls}(p_i)=\frac{1}{N_{com}}\sum\limits_{i\in\psi}F_i L_{fl}(p_i,g_i^*)\\
   &+ \frac{1}{N_{norm}}\sum\limits_{i\in\Omega}(1_{(l_i^*=0)}1_{(F_i<0.5)}+1_{(l_i^*=1)})L_{fl}(p_i,l_i^*)
\end{array}$
}
\end{equation}	

    where $i$ is the anchor index in a training-batch, $p_i$ is the predicted probability of the anchor $i$.  ${l_i}^*$ is the class label of anchor $i$, which is assigned {with} the label on the first step of standard anchor matching strategy. ${g_i}^*$ is the label of our newly compensated anchors, which are revised from backgrounds to foregrounds. $F_i$ is the IoU between the corresponding regression bounding box and its target ground-truth. $\Omega$ represents a set of all matched and unmatched low-quality anchors\footnote{This denotes the IoU between the predicted bounding box regressed by unmatched anchor and the target face is below 0.5.} and $\psi$ represents a set of newly compensated anchors. ${N_{norm}}$ is the number of {normally} matched anchors in $\Omega$ and $N_{com}$ is the total number of compensated anchors in $\psi$. $L_{fl}$ is the normal sigmoid focal loss over two classes (face foreground and background). In {addition}, {the supervision for new compensated anchors is added to the location loss} and the specific equation is shown as below:
\begin{equation}
\resizebox{0.43\textwidth}{!}{$
\begin{array}{ll}
L_{loc}(x_i)&=\frac{1}{N_{com}}\sum\limits_{i\in\psi}L_{SmoothL1}(x_i,x_i^*) \\
   &+\frac{1}{N_{norm}}\sum\limits_{i\in\Omega}L_{SmoothL1}(x_i,x_i^*)
\end{array}$
}
\vspace{-0.2cm}
\end{equation} 
    
where ${x_i}^*$ is the ground-truth location coordinates of anchor $i$. $L_{SmoothL1}$ is a normal location loss inspired by Faster-RCNN \cite{ren2015faster}. All other parameters are similar to $L_{cls}$'s.

\renewcommand{\algorithmicrequire}{ \textbf{Input:}}
\renewcommand{\algorithmicensure}{ \textbf{Output:}} 

\begin{algorithm}[htb]
\caption{Online high-quality anchor mining}
\label{algorithm_1}
\begin{algorithmic}[1]
\REQUIRE $B,G, T, K, D, L, R, A$\\
$B$ is a set of regression bounding boxes, in the form of ($x_0$, $y_0$, $x_1$, $y_1$). \\
$X$ is a set of ground-truth, in the form of ($x_0$, $y_0$, $x_1$, $y_1$) \\
$T$ is an online anchor mining threshold (see details on Subsection \ref{online_high_quality}) \\
$K$ is a hyperparameter and represents the max number of anchors that $F_{outer}$ can be matched with.  \\
$D$ is a Dict, key is ground-truth, item is the number of anchors that ground-truth can match in the first step of our HAMBox method. \\
$L$ is a Dict, key is anchor index, item is a label that anchor index is assigned with in the final process of our HAMBox method. \\
$R$ is a Dict, key is anchor index, item is encoded coordinates of the key during standard anchor matching strategy. \\
$A$ is a Dict, key is anchor index, item is coordinates of the key, in the form of ($x_0$, $y_0$, $x_1$, $y_1$). \\
\ENSURE $R$ and $L$ after using our HAMBox method.
\FOR{$x_i$ in $X$} 
\IF{$D(x_i) >= K$}
\STATE $continue$
\ENDIF
\STATE $CompensatedNumber=K - D(g_i)$ 
\STATE $OnlineIoU \Leftarrow {IoU(x_i,B), AnchorIdx}$
\STATE $SortedOnlineIoU = sorted(OnlineIoU, key=IoU, reverse=True)$
\FOR{$IoU,AnchorIdx$ in $SortedOnlineIoU$}
\IF{$L(AnchorIdx) = 1$}
\STATE $continue$
\ENDIF
\STATE $CompensatedNumber $ -= $ 1$
\STATE $L(AnchorIdx) = 1$
\STATE $R(AnchorIdx)=encode(A(AnchorIdx),ground$-$truth)$
\IF{$CompensatedNumber = 0$}
\STATE $break$
\ENDIF
\ENDFOR
\ENDFOR 
\STATE \textbf{Return} $R$, $L$

\end{algorithmic}
\end{algorithm}
\section{Experiments}
 In this section, we first show the effectiveness of our proposed strategies with comprehensive ablative experiments. Then with the final optimal model, our approach achieves state-of-the-art results on face detection benchmarks. 

\subsection{Ablation Study}
The WIDER FACE dataset is used in this ablation study. This dataset has 32,203 images with 393,703 labeled faces with huge variability in scales, occlusions and poses. Our networks are only trained on the training set and evaluated both on validation and test set. Average Precision (AP) score is used as the evaluation metric. 

\textbf{Data Augmentation} Our models are trained with following data augmentation strategies:
\begin{itemize}
    \item Color distort: Apply some photometric distortions similar to \cite{howard2013some}.
    \item Data anchor sampling: This method \cite{tang2018pyramidbox} resizes all train images through reshaping a random face in this image to a smaller size.
    \item Horizontal flip: After data-anchor-sampling, the cropped image patch is resized to 640 $\times$ 640 and horizontally flipped with a probability of 0.5.
\end{itemize}

\textbf{Baseline}.
We build an anchor-based detector with ResNet-50 guided by the RetinaNet as our baseline face detector. It differs from the original RetinaNet \cite{lin2017focal} in the following {four} aspects: Firstly, we set 6 anchors whose scales are from the set \{16, 32, 64, 128, 256, 512\}, and all anchors’ aspect ratios are set to 1:1. Secondly, we use LFPN \cite{tang2018pyramidbox} instead of FPN \cite{lin2017feature} for feature fusion since top two high-level features are extracted from regions with little context and may introduce noise for detecting small faces. Thirdly, we do not use deep head owing to two main factors. One is that the time cost of the training process is too high, the other is that our baseline is significantly higher than that of any other SOTA works on the WIDER FACE hard dataset. Finally, the threshold of IoU for matched anchors is changed to 0.35, and ignore-zone is not implemented. 

\textbf{Optimization Details}.
All models are initialized with the pre-trained weights of ResNet-50 and fine-tuned on WIDER FACE training set. Each training iteration contains seven images per GPU for a 4 NVIDIA Tesla V100 GPUs server. The initial learning rate is set to 0.01 and decreases to 0.001 after 110k iterations. All the models are trained for 140k iterations by synchronized SGD. The momentum and weight decay are set to 0.9 and 5$\times10^{-5}$, respectively.

\begin{figure*}[t]
\vspace{-0.5cm}
    \centering
    \subfigure[AFW]{
    \includegraphics[width=2in, height = 1.5in]{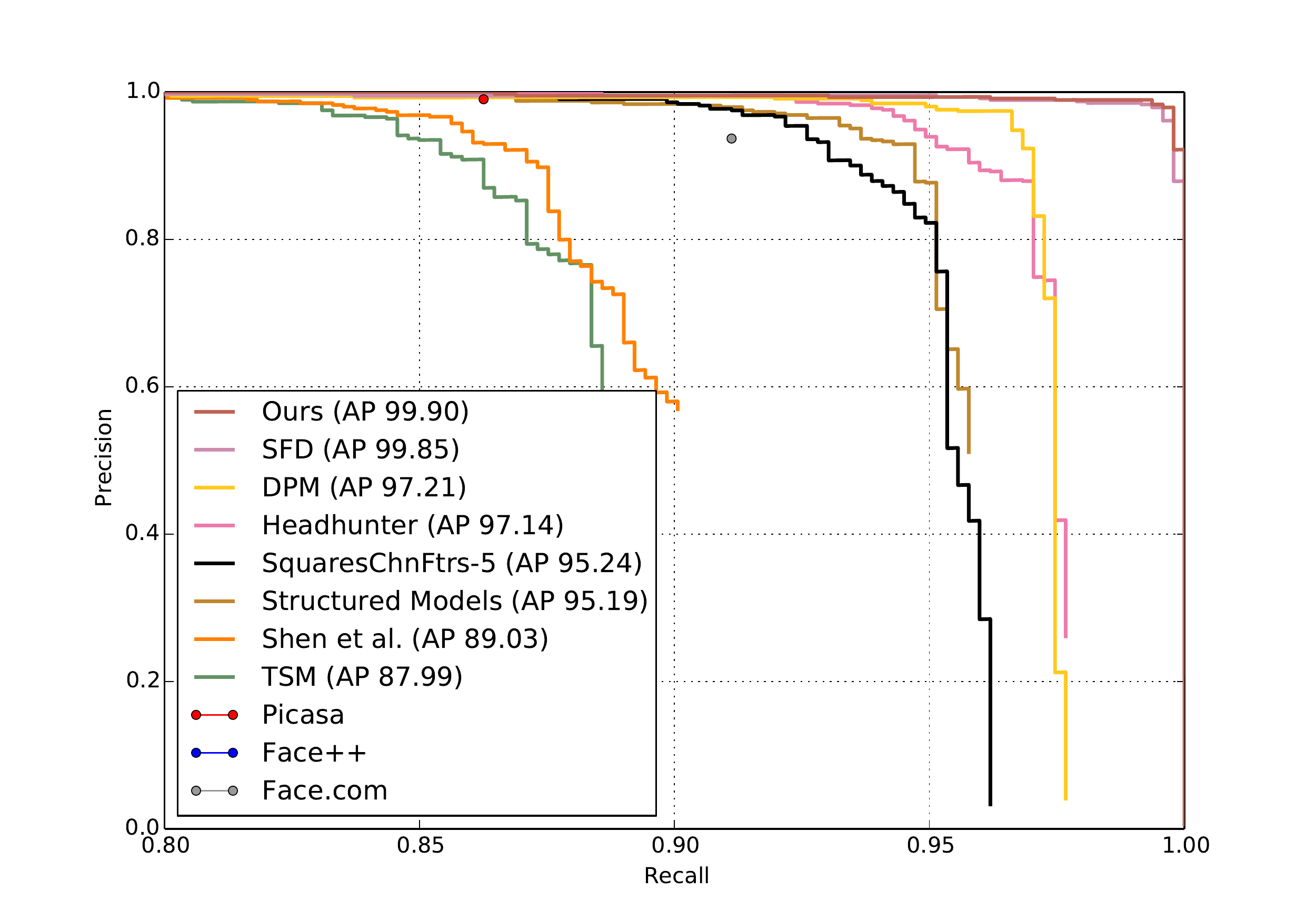}
    \label{img4_a}
    }
    \subfigure[PASCAL Face]{
    \includegraphics[width=2in, height = 1.5in]{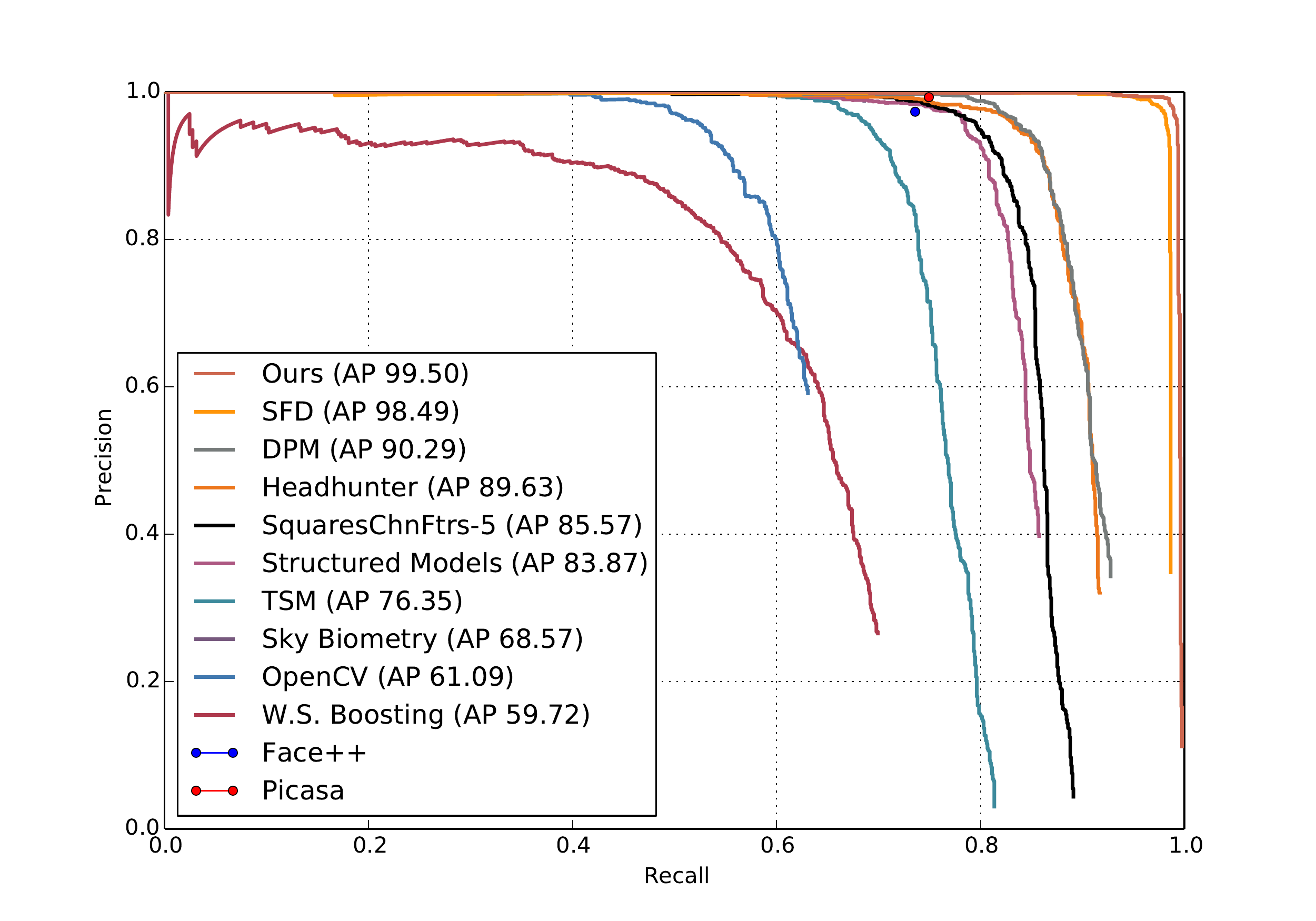}
    \label{img4_b}
    }
    \subfigure[FDDB]{
    \includegraphics[width=2in, height = 1.5in]{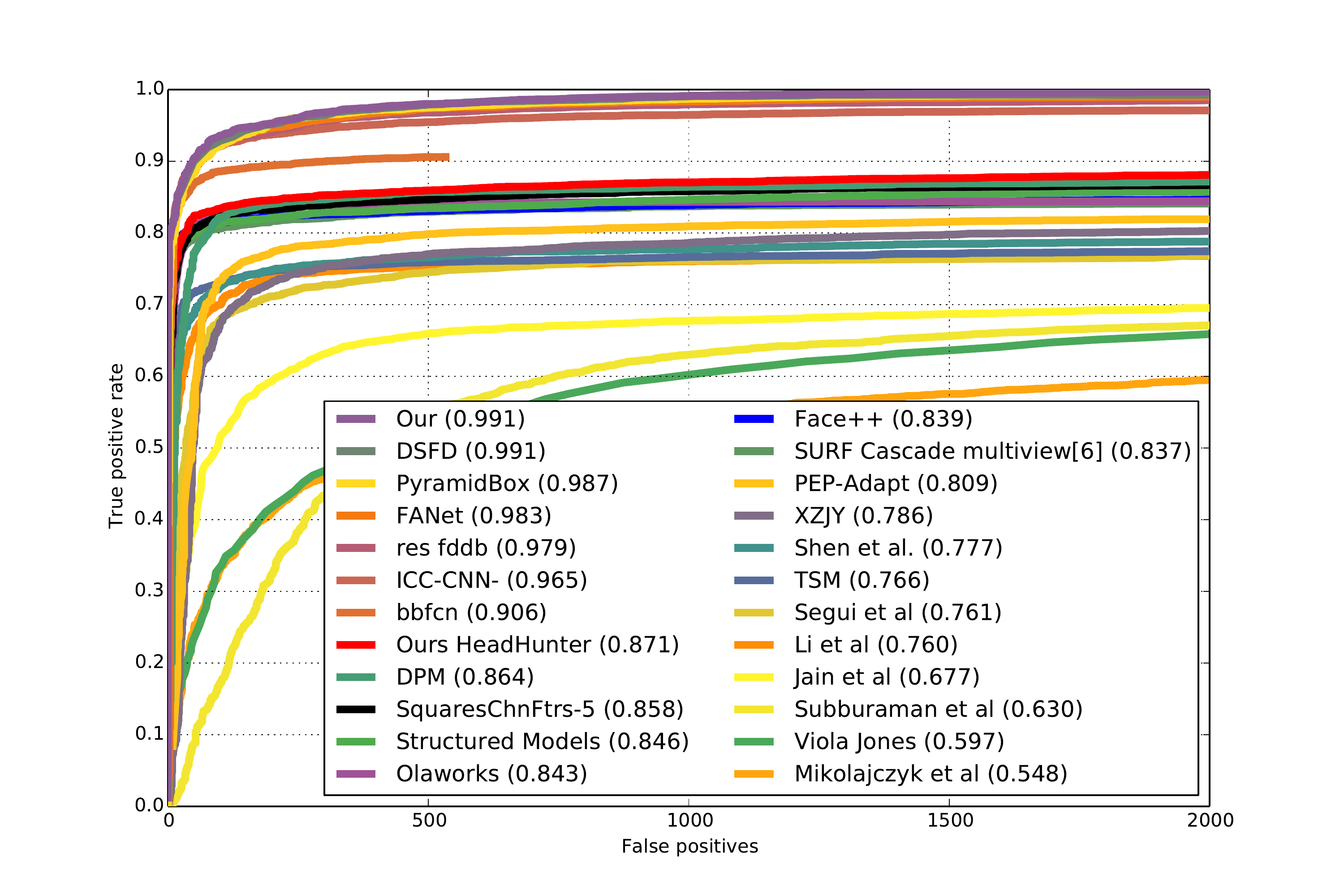}
    \label{img4_c}
    }
    \caption{Evaluation on the common face detection datasets.}
    \label{img_4}
\end{figure*}
\textbf{The effect of High-recall Anchor-based Detector}
As discussed above, the difference between our high-recall detector and baseline detector is the pre-defined anchor scale. Inspired by Figure \ref{img1}, we design our method with anchor scales \{16, 32, 64, 128, 256, 512\} * 0.68 tiled on pyramid feature maps from P2 to P6.

To better understand the advantage of our method, we conduct four experiments, which are shown in Table \ref{table_1}.
First step on standard anchor matching strategy with anchor scale ratio 0.68 (denoted as SMS({ratio=0.68})); 
Two-step on standard anchor matching strategy with anchor scale ratio 0.68 (denoted as DMS({ratio=0.68}));  
Two-step on standard anchor matching strategy with anchor scale ratio 0.5 whose scale could help more outer faces match anchor while significantly decreasing the number of anchors matched with each face (denoted as DMS({ratio=0.50}));  
New anchor matching strategy introduced by S$^3$FD \cite{zhang2017s3fd} with anchor scale ratio 0.68 (denoted as NAMS({ratio=0.68})). 
Compared to the baseline, SMS({ratio=0.68}), DMS({ratio=0.68}), NAMS({ratio=0.68}) provide a significant improvement on the hard subset (rising by 1.2\%, 0.8\%, 0.8\% AP respectively) and DMS({ratio=0.50}) is with no improvements on the hard subset (decreasing by 0.7\%). Through these experimental results, we could draw two conclusions: On the one hand, enhancing the proportion of faces that can be matched with anchors could improve the model performance. However, with the continuously decrease on the scale of anchor to enhance this proportion, the remaining faces are more difficult to match and the number of anchors that each face can match with decreases dramatically, which are the main reasons why the performance of DMS({ratio=0.50}) is 1.5\% AP lower than DMS({ratio=0.68}) on the hard dataset. On the other hand, NAMS({ratio=0.68}) and DMS({ratio=0.68}) achieve almost same performance with SMS({ratio=0.68}), suggesting that these two anchor compensation methods 
have less influence on the performance of the detector. Thus we use the SMS({ratio=0.68}) method in the following experiments. In addition, DMS({ratio=0.68}) and NAMS({ratio=0.68}) would be regarded as comparisons, respectively. {And the anchor ratio of SMS in Table \ref{table_3} and \ref{table_4} is set to 0.68.}

\begin{table}[h]
\scriptsize
\centering
\caption{AP performance on various anchor setting and anchor matching strategy on WIDER FACE validation subset.}
\label{table_1}

\setlength{\tabcolsep}{4mm}
\begin{tabular}{c|c|ccc}

\toprule
Subset  & {ratio} & Easy & Medium & Hard\\
\midrule
Baseline & 1.00 & 0.943 & 0.931 & 0.894 \\
\textbf{+SMS} & 0.68 & \textbf{0.949} & \textbf{0.945} & \textbf{0.906} \\
+DMS & 0.68 & 0.954 & 0.949 & 0.902 \\
+DMS & 0.50 & 0.938 & 0.922 & 0.887 \\
+NAMS & 0.68 & 0.951 & 0.948 & 0.902 \\
\bottomrule
\end{tabular}
\end{table}

\textbf{The Effect of Online {High-quality} Anchor Mining}
Next, we look into the effect of our proposed online high-quality anchor mining strategy. In this paragraph, we mainly discuss the effect of two hyper-parameters in our method. The performance under different $K$, $T$ (defined in Subsection \ref{online_high_quality}) is shown in Table \ref{table_2}. It shows that: (1) the performance gets better when $T$ increases and it is easier to {conclude} that the higher the quality of compensated anchors, the better the performance of the model. (2) The performance gets better when $K$ is smaller 
than 5 and gets worse when $K$ is larger than 5, suggesting that it is not good to increase too large numbers of compensated anchors that each outer face can match since the anchors {off-limits} are redundant for their corresponding faces. After multiple ablative experiments, we find the optimal K(3), T(8) and further increase the performance with 0.7\% AP.

\begin{table}[h]
\scriptsize
\centering
\caption{Varying $T$, $K$ for regression-aware focal loss on WIDER FACE validation subset.}
\label{table_2}

\setlength{\tabcolsep}{6mm}{
\begin{tabular}{c|c|ccc}

\toprule
$K$ & $T$   & Easy  & Medium & Hard  \\
\midrule
3 & 0.5 & 0.945 & 0.939  & 0.902 \\
7 & 0.5 & 0.941 & 0.937  & 0.898 \\
3 & 0.7 & 0.947 & 0.943  & 0.911 \\
5 & 0.7 & 0.952 & 0.941  & 0.909 \\
\textbf{3} & \textbf{0.8} & \textbf{0.957} & \textbf{0.951}  & \textbf{0.913} \\
3 & 0.9 & 0.962 & 0.943  & 0.911 \\
\bottomrule

\end{tabular}}
\end{table}

\textbf{The Effect of Regression-aware Focal Loss}
This regression-aware focal loss completes our final HAMBox model. As discussed above, this loss gives those new matched anchors a reasonable weight which simultaneously helps model training these anchors more steadily and precisely. Results using our regression-aware focal loss (denoted as RAL) are shown in Table \ref{table_4},  {and} the performance of our detector continues to increase 0.3\% AP.

\begin{figure*}[t]
    \centering
    \subfigure[Val: Easy]{
    \includegraphics[width=2in, height = 1.4in]{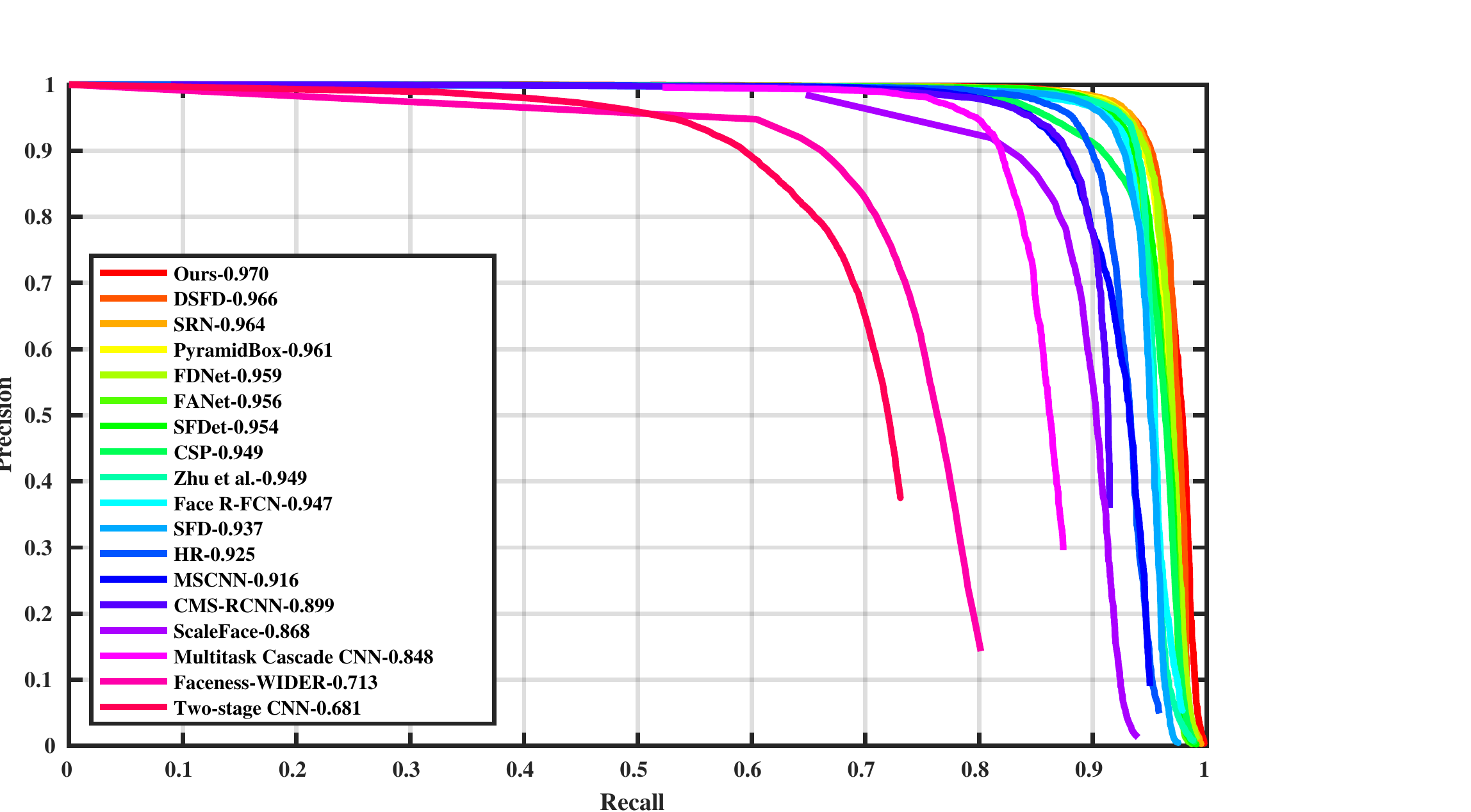}
    }
    \subfigure[Val: Medium]{
    \includegraphics[width=2in, height = 1.4in]{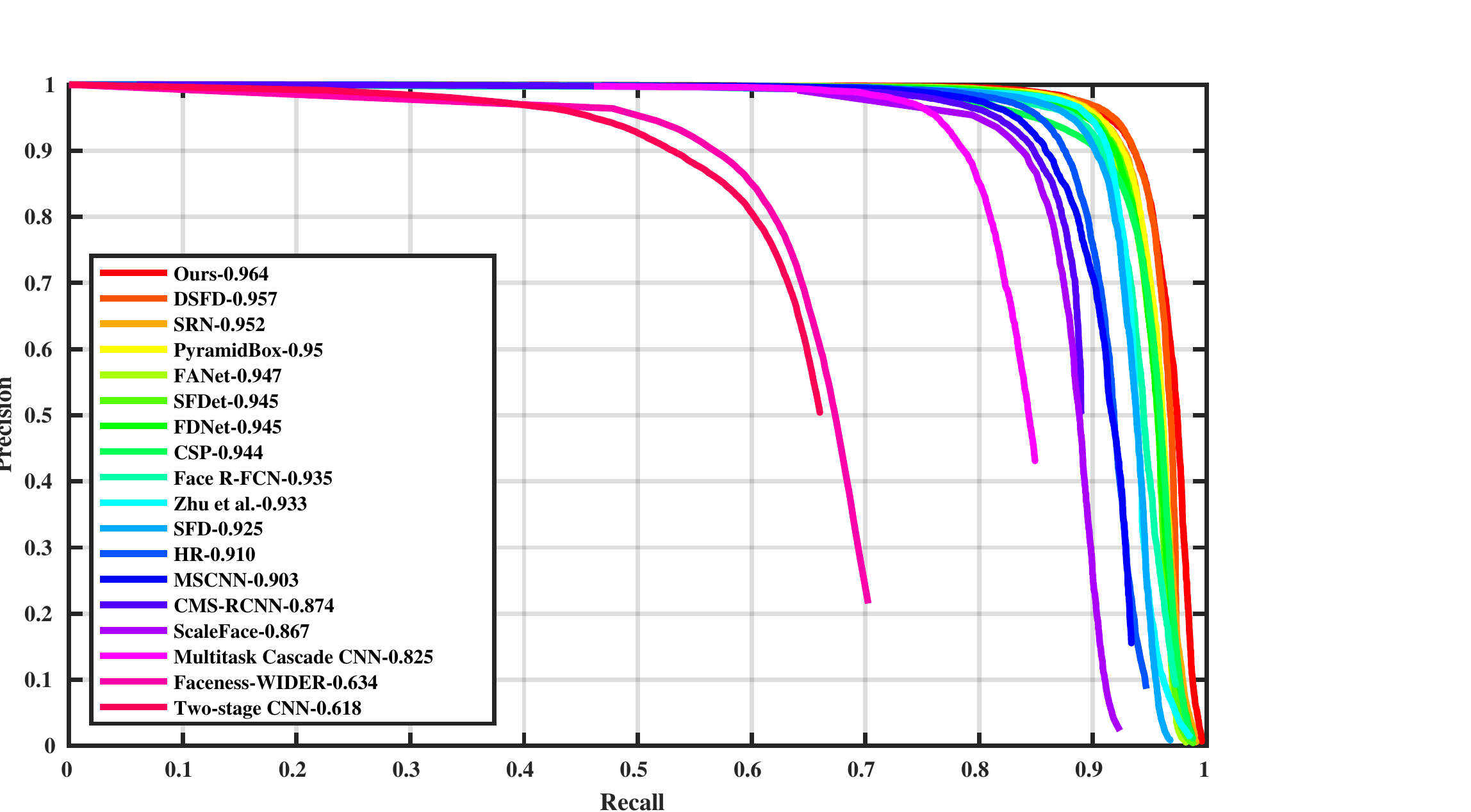}
    }
    \subfigure[Val: Hard]{
    \includegraphics[width=2in, height = 1.4in]{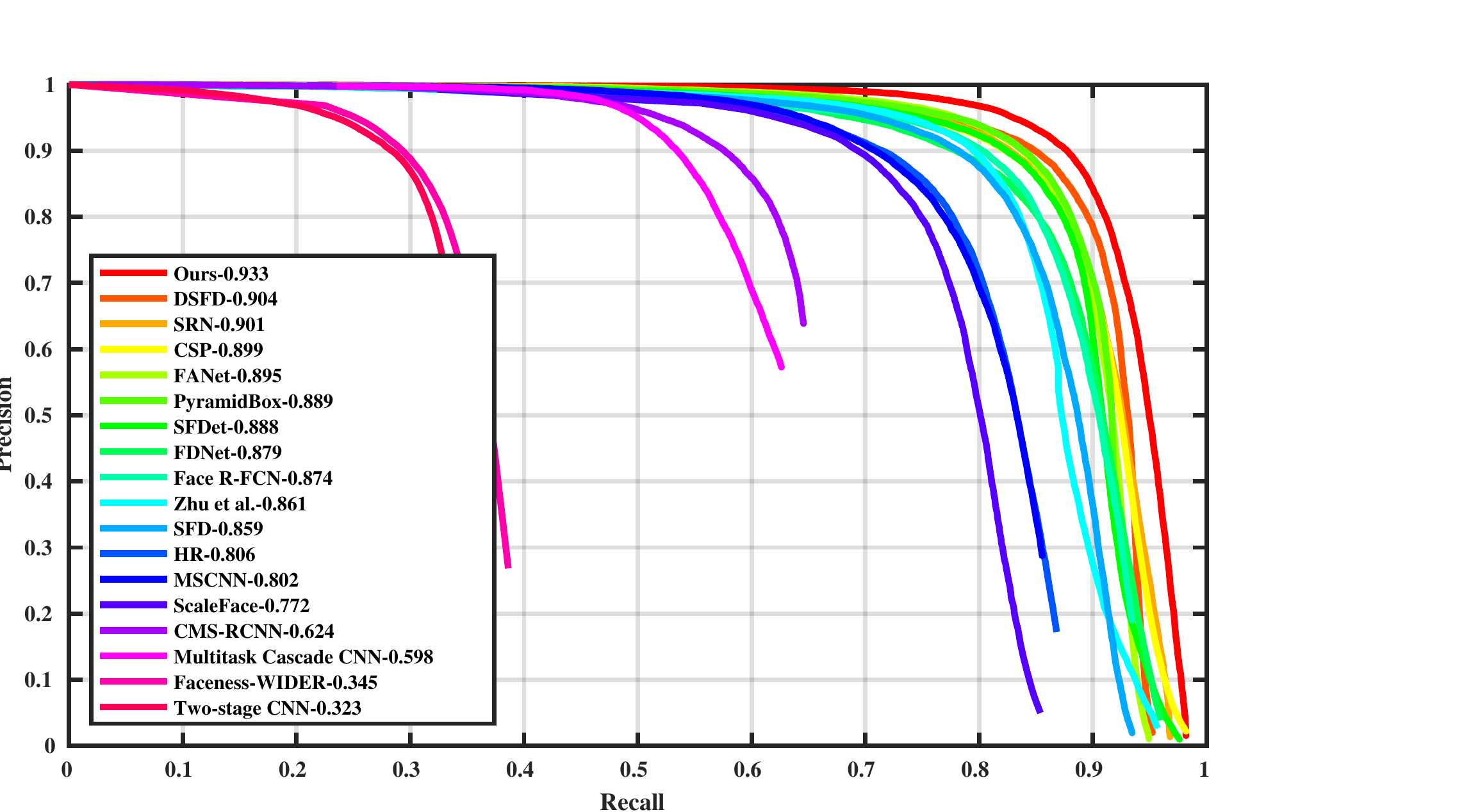}
    }
   	 
    \subfigure[Test: Easy]{
    \includegraphics[width=2in, height = 1.4in]{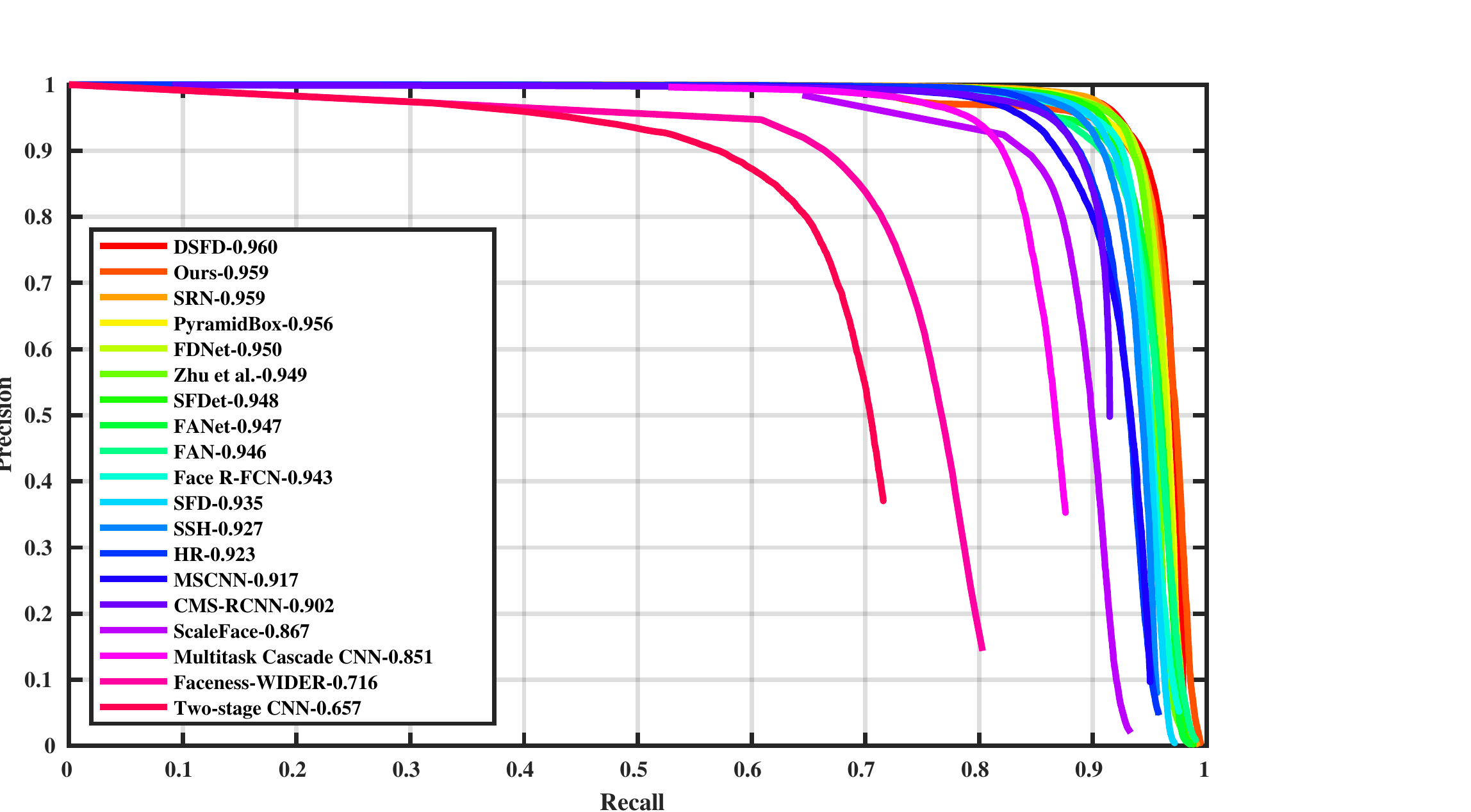}
    }
    \subfigure[Test: Medium]{
    \includegraphics[width=2in, height = 1.4in]{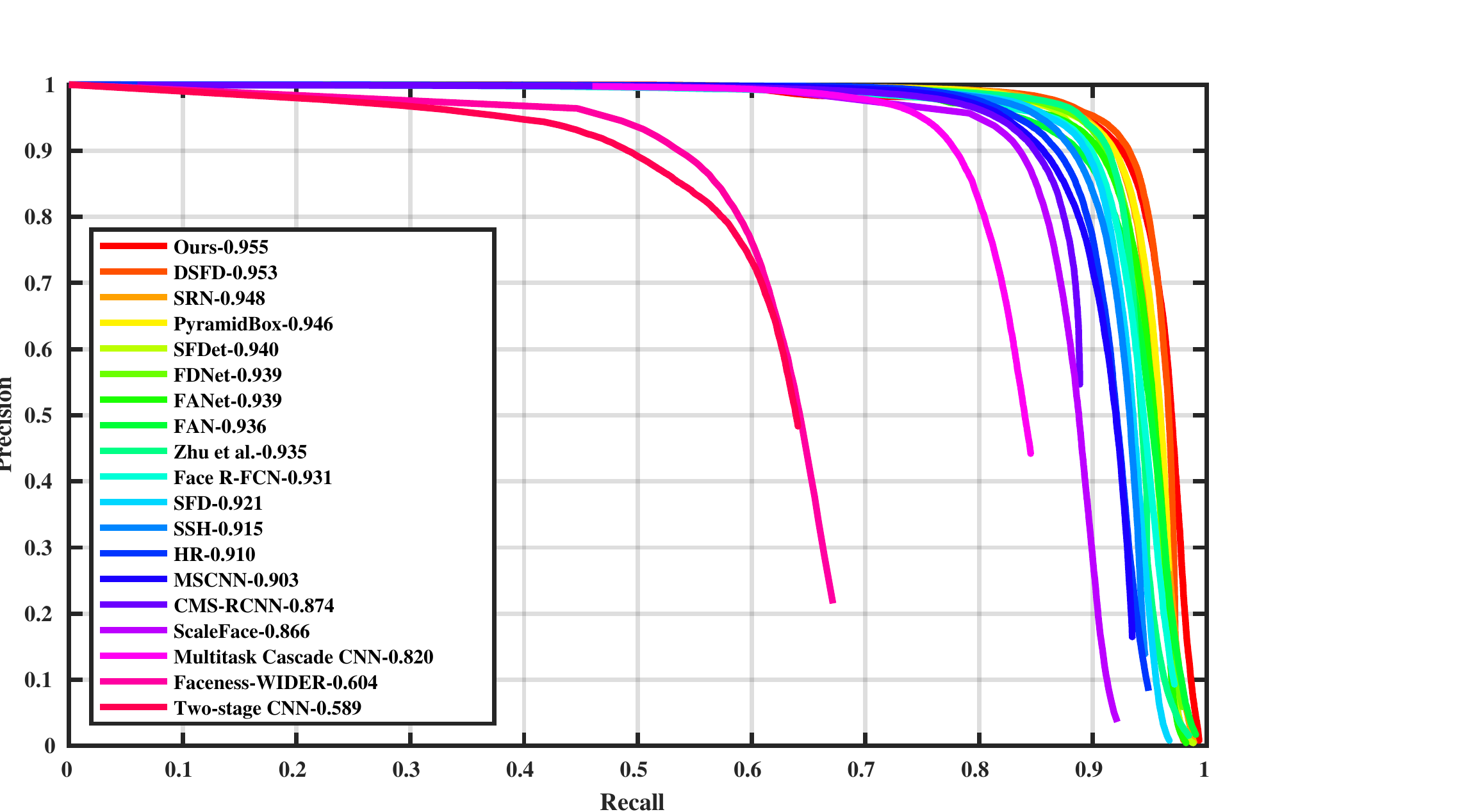}
    }
    \subfigure[Test: Hard]{
    \includegraphics[width=2in, height = 1.4in]{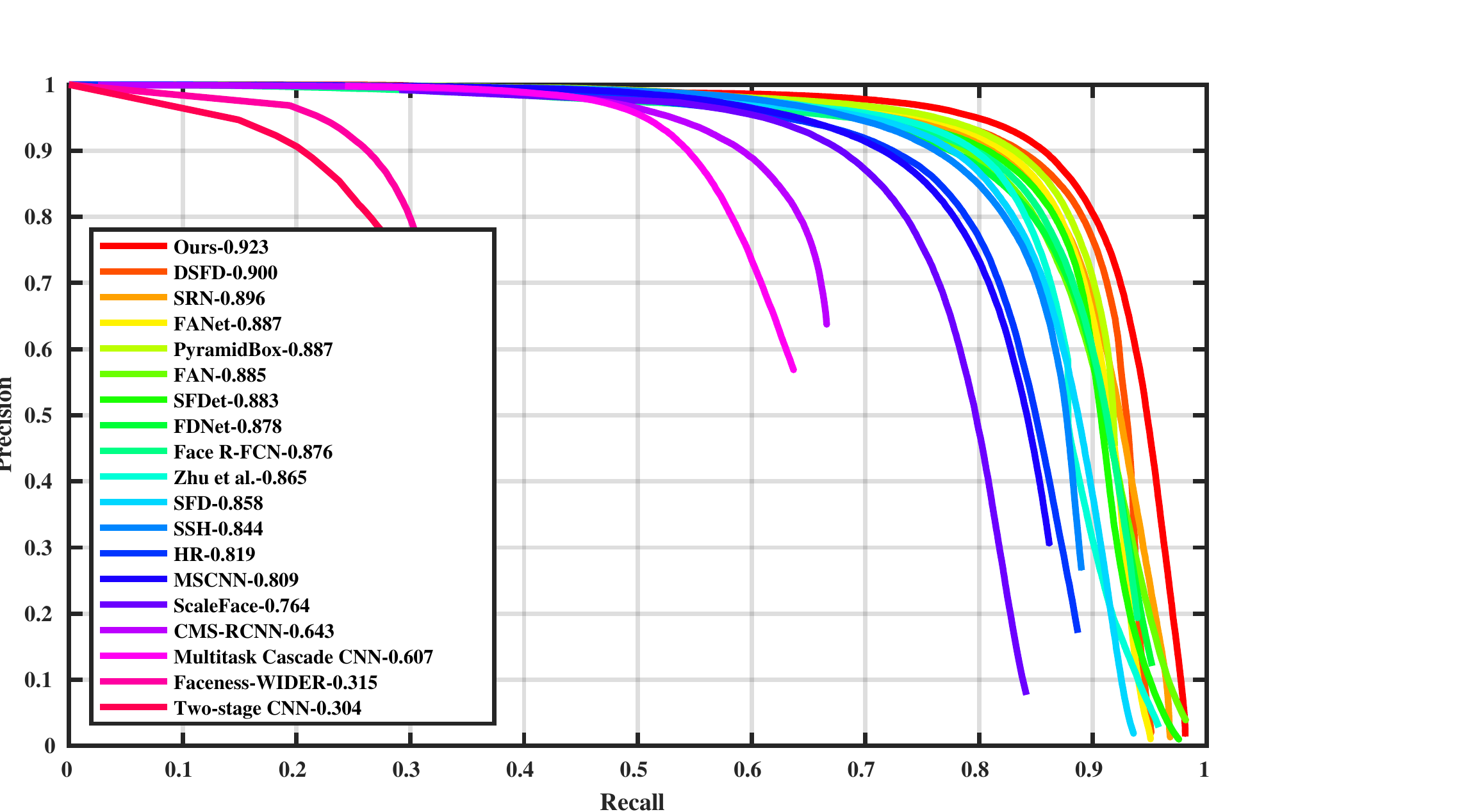}
    }
    \caption{Precision-Recall (PR) curves on WIDER FACE validation and testing subsets.}
    \label{img_5}
\end{figure*}
\textbf{Our method vs ZCC and NAMS}
\label{zcc_name}
To {further verify the effectiveness of our method}, we 
compare our method with NAMS \cite{zhang2017s3fd} and ZCC \cite{zhu2018seeing}. As shown in Table \ref{table_3}, our method outperforms theirs 6.4\% and 5.5\% AP respectively on their paper baseline. {Moreover,} in our baseline, ours also outperforms theirs 2.0\% and 3.4\% AP, respectively. Note that our method offers more high-quality {anchors} to help bounding box regression and classification branch optimize well.

\textbf{The Effect of Other Tricks}
As shown in Table \ref{table_4}, we introduce SSH \cite{najibi2017ssh}, deep head (DH) \cite{lin2017focal} and pyramid anchor (PA) \cite{tang2018pyramidbox} modules to further improve the performance of detector and achieve best AP among all state-of-the-art face detectors \cite{li2019dsfd,chi2019selective,liu2019high,tang2018pyramidbox,wang2017detecting,zhu2018seeing,zhang2017s3fd,najibi2017ssh,hu2017finding}. We outperform others on validation/ test hard dataset 2.9\%, 2.3\% AP, respectively. 
Besides, we achieve 57.45\%, 57.13\% on validation/ test datasets when using more scientific mAP score metric.
\begin{table}[h]
\scriptsize
\centering
\caption{AP performance of our model with various anchor matching strategy on WIDER FACE validation subset. * denotes the reproduced performance by us, and APs in NAMS and ZCC represent the performance presented by their papers.}
\label{table_3}

\setlength{\tabcolsep}{1mm}

\begin{tabular}{c|ccc}
\toprule
Subset  &Easy & Medium       & Hard \\
 \midrule
NAMS                               & 0.937 & 0.924 & 0.852 \\
ZCC                              & 0.949 & 0.933 & 0.861 \\
Baseline + $NAMS^*$                               & 0.941 & 0.937 & 0.896 \\
Baseline  + $ZCC^*$                              & 0.943 & 0.942 & 0.882 \\
\textbf{Baseline  + {SMS} + OAM + RAL} & \textbf{0.962}        & \textbf{0.953}        & \textbf{0.916}        \\
\bottomrule
\end{tabular}
\end{table}

\begin{table}[h]
\scriptsize
\centering
\caption{AP performance of our proposed modules and {additional} tricks on WIDER FACE validation subset.}
\label{table_4}

\setlength{\tabcolsep}{1mm}

\begin{tabular}{ccccccc|ccc}
\toprule
Baseline & {SMS} & OAM & RAL & DH & SSH & PA & Easy & Medium & Hard\\
\midrule
$\surd$ & - & - & - & - & - & - & 0.943  & 0.931 & 0.894 \\
$\surd$ & $\surd$ & - & - & - & - & - & 0.949  & 0.945 & 0.906 \\
$\surd$ & $\surd$ & $\surd$ & - & - & - & - & 0.957  & 0.951 & 0.913 \\
$\surd$ & $\surd$ & $\surd$ & $\surd$ & - & - & - & 0.962  & 0.953 & 0.916 \\
$\surd$ & $\surd$ & $\surd$ & $\surd$ & $\surd$ & - & - & 0.964  & 0.955 & 0.922 \\
$\surd$ & $\surd$ & $\surd$ & $\surd$ & $\surd$ & $\surd$ & - & 0.968  & 0.959 & 0.927 \\
$\surd$ & $\surd$ & $\surd$ & $\surd$ & $\surd$ & $\surd$ & $\surd$ & \textbf{0.970} & \textbf{0.964} & \textbf{0.933} \\
\bottomrule
\vspace{-0.5cm}
\end{tabular}
\end{table}

\subsection{Evaluation on Common Benchmarks}
We evaluate our proposed method on the common face detection benchmarks, including WIDER FACE \cite{yang2016wider}, Annotated Faces in the Wild (AFW) \cite{zhu2012face}, PASCAL Faces \cite{yan2014face}, FDDB \cite{jain2010fddb}. Our face detector is trained only using WIDER FACE training set and is tested on those benchmarks. We demonstrate the state-of-the-art performance across all the datasets.  

\textbf{WIDER FACE Dataset} We report the performance of our face detection system on the WIDER FACE \cite{yang2016wider} testing set with 16,097 images. Detection results are sent to the database server for receiving the precision-recall curves. Figure \ref{img_5} illustrates the precision-recall curves along with AP scores. Our proposed method achieves 97.0\% (Easy), 96.4\%(Medium), 93.3\%(Hard) on validation dataset and 95.9\% (Easy), 95.5\% (Medium), 92.3\% (Hard) on test dataset. Especially on the hard subset, we outperform the current state-of-the-art model 2.3\% AP (Test) and 2.9\% AP (Validation). This huge enhancement demonstrates the superiority of our method.

\textbf{AFW Dataset} This dataset \cite{zhu2012face} consists of 205 images with 473 annotated faces. 
Figure \ref{img4_a} shows that our detector outperforms others by a considerable margin. 

\textbf{PASCAL Face Dataset} This dataset \cite{yan2014face} 
has 851 images with 1,335 annotated faces. 
Figure \ref{img4_b} demonstrates the superiority of our method. 

\textbf{FDDB Dataset} This dataset \cite{jain2010fddb} has 2,845 images with 5,171 annotated faces. Most of them are with low image resolutions and complicated scenes, such as occlusions, huge poses. Figure \ref{img4_c} shows our {proposed} method outperforms all state-of-the-art models.
\vspace{-0.2cm}

\section{Conclusion}
\vspace{-0.1cm}
In this paper, we first observe an interesting phenomenon that only 11\% correctly predicted bounding boxes are regressed from the unmatched anchors in the inference phase. Then we further propose an online high-quality anchor mining strategy that helps outer faces match high-quality anchors. Our method first enhances the proportion of face matched with anchor, and then we propose an online high-quality anchor compensation strategy for outer faces. Finally, we design a regression-aware focal loss for new compensated anchors. We conduct extensive experiments on the AFW, PASCAL Face, FDDB, WIDER FACE datasets and achieve the state-of-the-art detection performance.

{\small
\bibliographystyle{ieee_fullname}
\bibliography{ref}
}

\end{document}